%% file: main.tex
\documentclass[letterpaper, 10 pt, journal, twoside]{IEEEtran}

\IEEEoverridecommandlockouts                              

\usepackage{amsmath}
\usepackage{mathtools,amssymb}
\usepackage{hyperref}
\usepackage{bm}
\usepackage{color}
\usepackage{xcolor}
\usepackage{cite}
\usepackage{booktabs}
\usepackage{multirow}
\usepackage[linesnumbered,ruled,vlined]{algorithm2e}
\usepackage[binary-units]{siunitx}
\usepackage{subcaption}
\usepackage{makecell}
\usepackage{graphicx}
\usepackage{balance}
\usepackage{algorithmic}
\usepackage{gensymb}
\usepackage{lipsum}  
\usepackage{soul}
\usepackage{siunitx}

\SetKwComment{Comment}{/* }{ */}
\SetKwInput{KwInput}{Input}                
\SetKwInput{KwOutput}{Out}              
\SetKwFor{KwFor}{for each}{do}{}
\SetKwInput{KwData}{Global params.}

\usepackage{mwe,tikz}
\usetikzlibrary{backgrounds,arrows,automata,shapes,positioning,calc,through,math}
\tikzset{
  imgletter/.style={
    rectangle,
    inner sep=2pt,
    rounded corners=.1em,
    text=black,
    minimum height=1em,
    text centered,
    fill=white,
    fill opacity=.7,
    text opacity=1,
    anchor=south west,
  },
}

\newtheorem{theorem}{Theorem}
\newtheorem{definition}[theorem]{\textbf{Definition}}

\newcommand{\pmin}[0]{P_{\mathrm{min}}^{ij}}
\newcommand{\pmax}[0]{P_{\mathrm{max}}^{ij}}

\newcommand{\disable}[1]{}

\newcommand{\changed}[1]{{\color{black}#1}}
\newcommand{\edited}[1]{{\color{black}#1}}

\usepackage{eso-pic} 
\usepackage{pdfpages}
\newcommand{\placetextbox}[3]{
  \setbox0=\hbox{#3}
  \AddToShipoutPictureFG*{
    \put(\LenToUnit{#1\paperwidth},\LenToUnit{#2\paperheight}){\vtop{{\null}\makebox[0pt][c]{#3}}}%
  }%
}%

\title{
\vspace{0.4em}\huge CTopPRM: Clustering Topological PRM for Planning Multiple Distinct Paths in 3D Environments
}

\author{Matej Novosad, Robert Penicka, Vojtech Vonasek
\thanks{\edited{Manuscript received: May 25, 2023; Revised August 8, 2023; Accepted September 6, 2023.}}
\thanks{\edited{This paper was recommended for publication by Editor Chao-Bo Yan upon evaluation of the Associate Editor and Reviewers' comments.}}
\thanks{
The authors are with the Multi-robot Systems Group, Faculty of Electrical
Engineering, Czech Technical University in Prague, Czech Republic (\protect\url{http://mrs.felk.cvut.cz/}).
This work  has been supported by the Czech Science Foundation (GAČR) under research project No. 22-24425S, by European Union’s Horizon 2020 research and innovation programme AERIAL-CORE under grant agreement no. 871479, and by CTU grant no SGS23/177/OHK3/3T/13.%
}
\thanks{\edited{Digital Object Identifier (DOI): 10.1109/LRA.2023.3315539.}}
}

\renewcommand\subsubsection[1]{\vspace{0pt}\noindent\textbf{#1.}}

\begin{document}

\placetextbox{0.5}{0.956}{
\fbox{
\begin{minipage}{\dimexpr\textwidth-2\fboxsep-2\fboxrule\relax}
M. Novosad, R. Penicka and V. Vonasek, \textbf{CTopPRM: Clustering Topological PRM for Planning Multiple Distinct Paths in 3D Environments}, in IEEE Robotics and Automation Letters, \url{https://doi.org/10.1109/LRA.2023.3315539}.
\end{minipage}
}
}%

\setlength{\abovedisplayskip}{6pt}
\setlength{\belowdisplayskip}{6pt}
\setlength{\abovedisplayshortskip}{4pt}
\setlength{\belowdisplayshortskip}{4pt}

\maketitle

\begin{abstract}
In this paper, we propose a new method called Clustering Topological PRM (CTopPRM) for finding multiple \changed{topologically} distinct paths in 3D cluttered environments.
Finding such distinct paths, e.g., going around an obstacle from a different side, is useful in many applications.
Among others, it is necessary for optimization-based trajectory planners where found trajectories are restricted to only a single \changed{topological} class of a given path.
Distinct paths can also be used to guide sampling-based motion planning and thus increase the effectiveness of planning in environments with narrow passages.
Graph-based representation called roadmap is a common representation for path planning and also for finding multiple distinct paths.
However, challenging environments with multiple narrow passages require a densely sampled roadmap to capture the connectivity of the environment.
Searching such a dense roadmap for multiple paths is computationally too expensive. 
Therefore, the majority of existing methods construct only a sparse roadmap which, however, struggles to find all distinct paths in challenging environments.
To this end, we propose the CTopPRM which creates a sparse graph by clustering an initially sampled dense roadmap. 
Such a reduced roadmap allows fast identification of \changed{topologically} distinct paths captured in the dense roadmap.
We show, that compared to the existing methods the CTopPRM improves the probability of finding all distinct paths by almost 20\% in tested environments, during same run-time. 
The source code of our method is released as an open-source package.
\end{abstract}
\begin{IEEEkeywords}
Motion and Path Planning; Planning, Scheduling and Coordination
\end{IEEEkeywords}


\vspace{-0.7em}
\section*{Supplementary Material}
{\small
\vspace{-0.3em}
\noindent \textbf{Video:} \url{https://youtu.be/azNrWBU5cAk}\\
\noindent \textbf{Code:} \url{https://github.com/ctu-mrs/CTopPRM}
\vspace{-0.7em}
}

\section{Introduction\label{sec:intro}}

\begin{figure}[t]
     \centering
      \begin{tikzpicture}
          \node[anchor=south west,inner sep=0] (img) at (0,0) {\includegraphics[width=0.49\columnwidth]{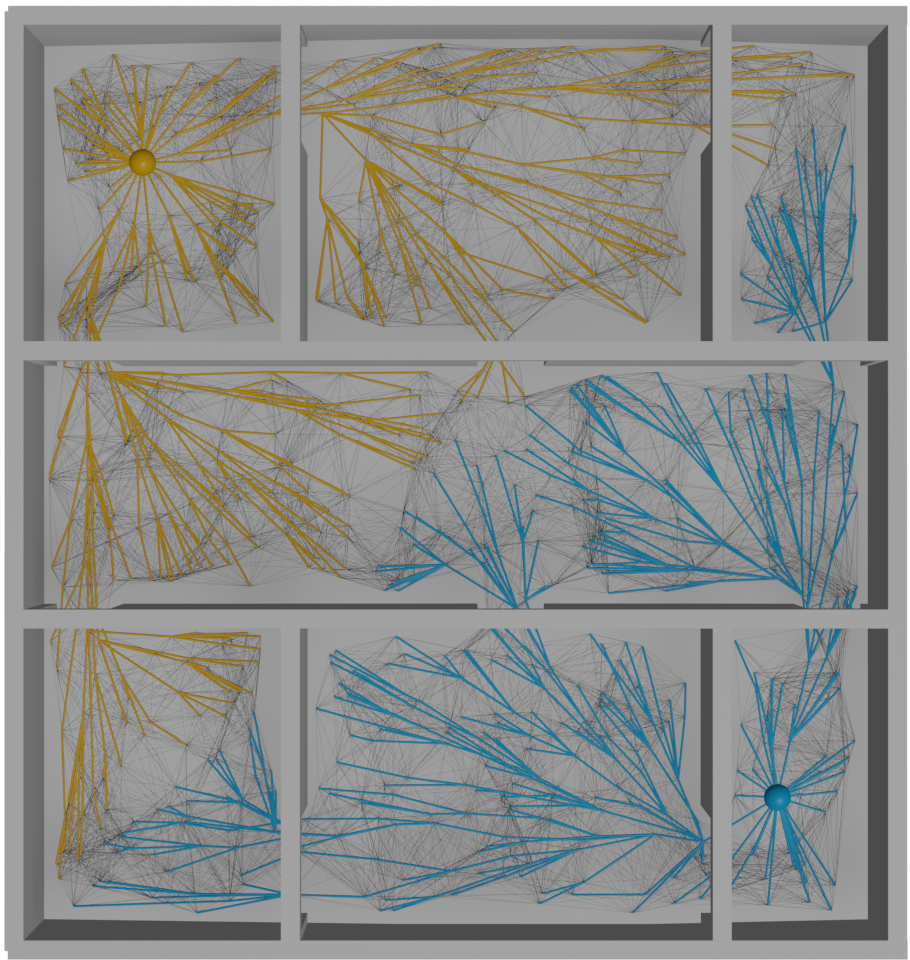}};
          \node[imgletter,text=black] (label) at (img.south west) {(a)};
      \end{tikzpicture}
      \begin{tikzpicture}
        \node[anchor=south west,inner sep=0] (img) at (0,0) {\includegraphics[width=0.49\columnwidth]{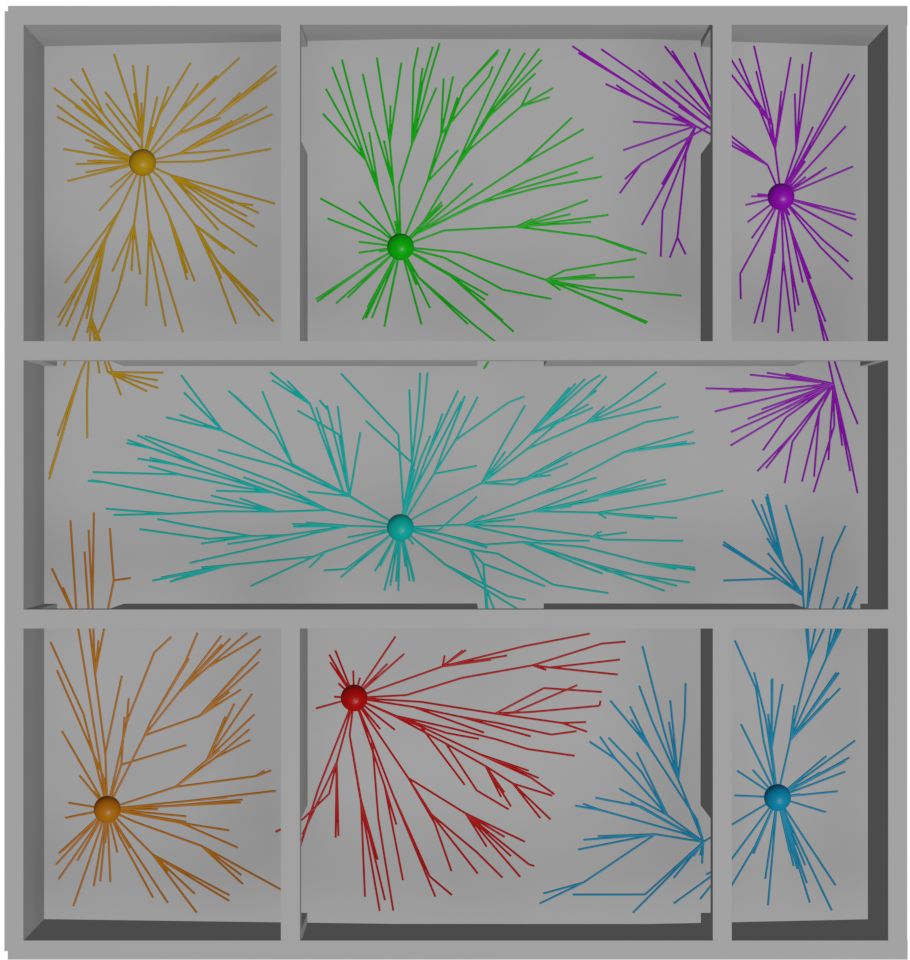}};
        \node[imgletter,text=black] (label) at (img.south west) {(b)};
      \end{tikzpicture}
      \begin{tikzpicture}
        \node[anchor=south west,inner sep=0] (img) at (0,0) {\includegraphics[width=0.49\columnwidth]{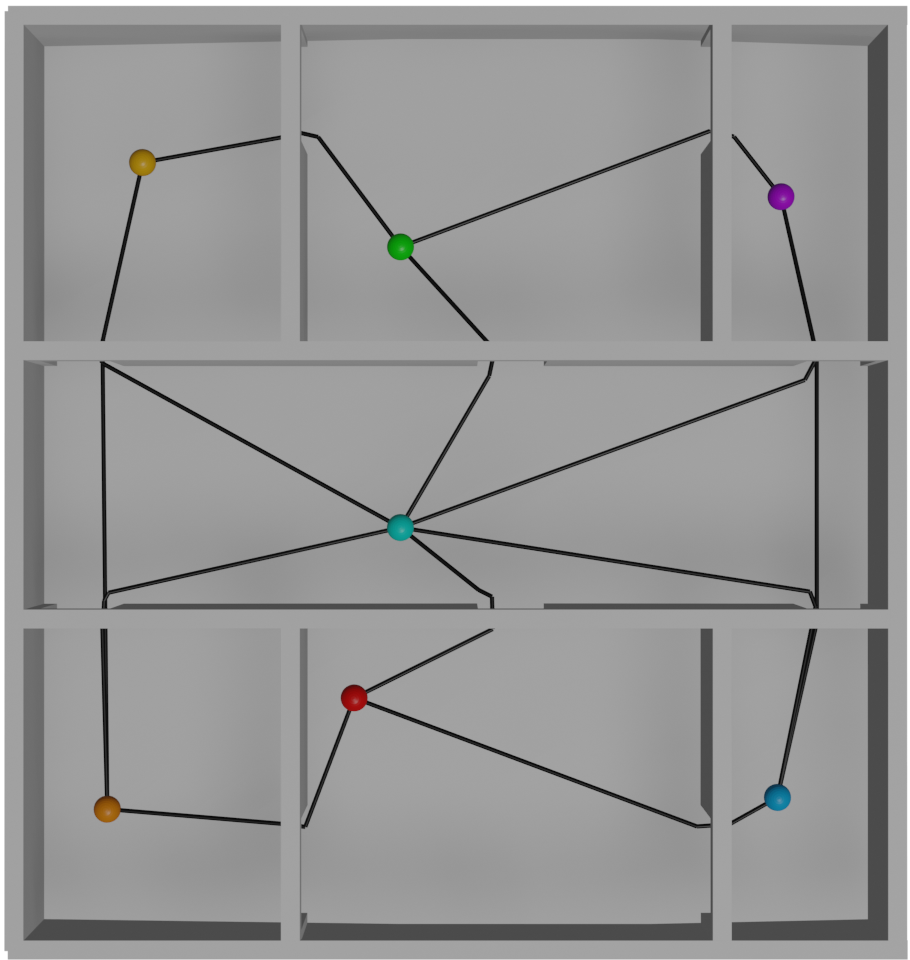}};
        \node[imgletter,text=black] (label) at (img.south west) {(c)};
      \end{tikzpicture}
      \begin{tikzpicture}
        \node[anchor=south west,inner sep=0] (img) at (0,0) {\includegraphics[width=0.49\columnwidth]{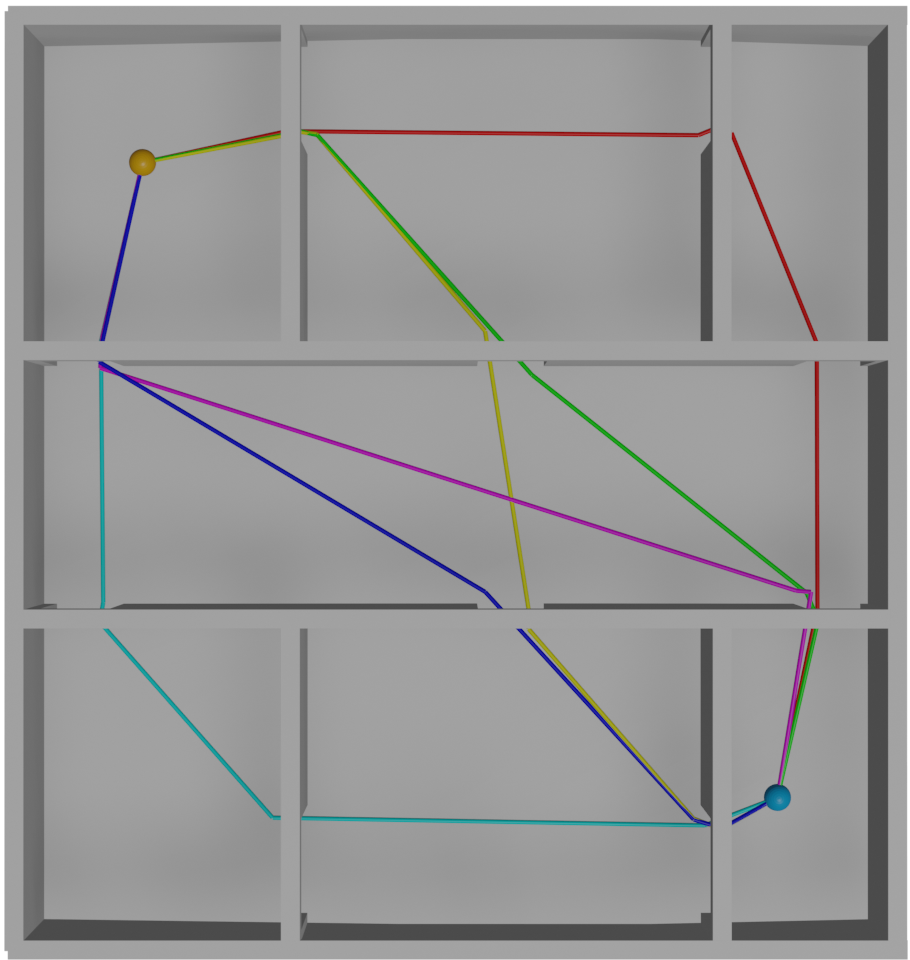}};
        \node[imgletter,text=black] (label) at (img.south west) {(d)};
      \end{tikzpicture}
        \vspace{-1.3em}
        \caption{
        Illustration of the proposed CTopPRM method which starts by creating a dense Probabilistic Roadmap clustered around the start and goal positions (a). 
        New cluster centroids are then iteratively added to promising places in the roadmap (b) to create a sparse graph (c), which is finally used to find distinct paths (d). 
        \label{fig:illustration}}
        \vspace{-1.7em}
\end{figure}
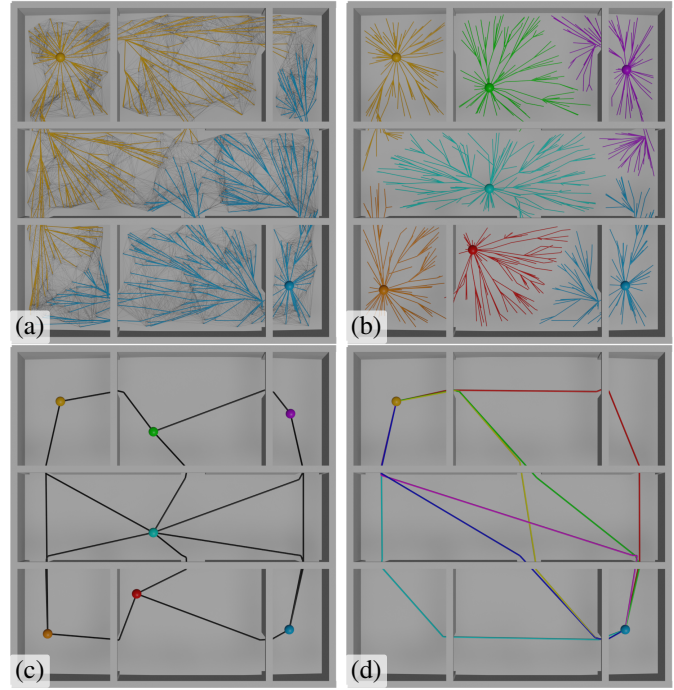

\IEEEPARstart{P}{ath} planning~\cite{lavalle2006planning} is one of the fundamental problems in robotics. 
It requires finding a geometrical path for a robot between given start and goal positions while avoiding collisions. 
However, there are several applications that would benefit from having multiple alternative paths.
\changed{To address this issue, topologically distinct paths, representing the topological connectivity of a cluttered environment  should be considered. }
\changed{Multiple paths allow the robot to select different ways how to navigate through the environment, see Fig.~\ref{fig:illustration}(d) with multiple distinct paths in a building-like environment.}

Finding multiple paths with distinct \changed{topological} classes can be used, for example, in optimization-based~\cite{zhou2021raptor} or sampling-based~\cite{penicka2022quadrotor} trajectory planners that have to consider robot dynamics.
Finding multiple paths helps to find the optimal trajectory as the trajectory planning is restricted to a single \changed{topological} class of a given initial path. 
Similarly, the paths can also be used to guide reinforcement learning methods for agile flight~\cite{penicka2022RL}. 
Last but not least, finding multiple distinct paths is beneficial for guided-based planners solving high-dimensional motion planning problems~\cite{denny2016theory, vonasek2019iros, denny2020dynamic, vonasek2020searching, belter2022walking}. 
These planners sample the configuration space of the robot
 around the guiding paths and thus increase the effectiveness of planning in environments with narrow passages.
However, while online trajectory planning requires fast computation, guided-based planners and reinforcement learning methods often benefit more from acquiring a higher number of guiding paths. 
Therefore, when searching for multiple \changed{topologically} distinct paths, a trade-off between computational time and the number of found distinct paths has to be considered.
This was so far the main stumbling block of existing methods.

The majority of existing methods for planning multiple distinct paths use a graph-based roadmap representation.
To discover all distinct paths, especially in challenging 3D environments, a dense roadmap is required.
However, searching for multiple paths as proposed in~\cite{fujita2003dualdijkstra} proves to be computationally expensive, even more so in a dense roadmap. 
Moreover, many found paths would belong to the same \changed{topological} class, requiring an exhaustive filtering process. 
Visibility-PRM~\cite{simeon2000visibilityprm} introduced a concept that allows the construction of a sparse roadmap, that was used by~\cite{zhou2021raptor},~\cite{simeon2008deformation} and~\cite{schmitzberger2002hppr}. 
Yet, Visibility-PRM's reliability, i.e. the ability to consistently capture all \changed{topological} classes, is limited, particularly in environments with narrow passages\changed{, which require a higher density of samples to capture paths leading through them}. 
However, finding multiple distinct paths in the dense roadmaps is computationally very demanding. 

To this end, we propose a novel sampling-based method called Clustering Topological PRM (CTopPRM), that clusters a dense roadmap, to construct a sparse graph with cluster centroids as vertices.
This reduced roadmap allows fast path searching while capturing all \changed{topological} classes that the initial dense roadmap had captured, including those that require traversal of narrow passages. 
\changed{Moreover, the algorithm enables adjusting the trade-off between computational time and number of paths found using tunable parameters.}
\changed{This makes it suitable for both online planning within tens of milliseconds, and for offline planning with narrow passages.}

We consider the contributions of this paper to be as follows. 
We introduce an efficient method for identifying \changed{topologically} distinct paths, with a controllable balance between computational time and the quantity of identified paths. 
We demonstrate that our approach, called CTopPRM, outperforms existing methods in a variety of challenging cluttered environments.
In scenarios with a \changed{low} number of distinct \changed{topological} classes, CTopPRM is shown to successfully identify \SI{94}{\percent} of all distinct paths while other methods find less than \SI{80}{\percent} of paths within the same run-time.
In \changed{environments} with a high number of distinct \changed{topological} classes, we improved the average number of \changed{topological} classes detected within the same computational time between \SI{30}{\percent} and \SI{300}{\percent}. 
\changed{Finally, we release CTopPRM's source code and our implementations of related methods, along with testing environments, as an open-source package.}

\section{Related Work\label{sec:related}}

A complete solution to the problem of finding all \changed{topologically} distinct paths in cluttered environments relies on combinatorial motion planning approaches. 
However, these methods~\cite{rosman2017dfs, kuderer2014onlinegeneration} use representations (e.g., Voronoi diagram) that require an explicit representation of occupied space.
An optimization-based approach described in~\cite{huang2017gaussian} proposes to use Gaussian processes to construct a factor graph representing a distribution of multiple trajectories, which are then optimized and filtered. 
However, the functionality of this method was verified only in 2D. 
Additionally, some of the trajectories found belong to the same \changed{topological} class. 
Therefore, the resulting paths must be pruned by identifying \changed{topologically} equivalent paths.
Authors in~\cite{bhattacharya2012constraints} introduce homotopy relation in the form of h-signatures, applicable in both 2D and 3D, but only with time and memory-consuming space discretization.
Moreover, the discretized space often fails to capture narrow passages.


To approximate the continuous configuration space, a graph-based representation called roadmap, e.g., Probabilistic Roadmap (PRM)~\cite{kavraki1996PRM}, is commonly used. 
Many existing methods for finding distinct \changed{topological} classes~\cite{simeon2008deformation, schmitzberger2002hppr, zhou2021raptor, penicka2022quadrotor, zhang2019sparse, rosmasnn2015multiple} take one of the PRM variants as a starting point. 
The Probabilistic Roadmap algorithm is a widely used sampling-based method for motion planning that consists of two main phases. 
In the construction phase, the PRM algorithm generates a set of random feasible configurations, also known as samples. 
These samples are then connected to each other using a local planner. 

The original PRM implementation in~\cite{kavraki1996PRM} did not allow cycles in the roadmap, which limited its connectivity, completeness and the ability to capture more than one \changed{topological} class. 
To address this,~\cite{karmaman2011sPRM} introduced a version called sPRM that allows cycles in the graph, and is more widely used nowadays. 
The author of Informed PRM~\cite{aria2021informedprm} proposes to only sample an ellipsoid space between start and goal configurations. 
Method in~\cite{kala2016hrm} aims to generate PRM that guarantees to capture all \changed{topological} classes in an environment, by using an obstacle biased sampler, but relies on explicit representation of occupied space, which is unreasonable for 3D environments. 

After PRM is constructed, query phase follows where a path between two samples is found using any standard graph searching algorithm such as Dijkstra's or A*. 
However, these algorithms only find the shortest path in the roadmap. 
Method~\cite{fujita2003dualdijkstra} proposes an approach that uses Dijkstra's algorithm to find all paths between start and goal node by finding a path to each node from start and from goal, resulting in total number of paths equal to number of nodes.
Method then proceeds to prune any redundant paths, which is an exhaustive process, especially in dense graphs with a high number of nodes, which are necessary in challenging environments that contain multiple narrow passages. 
Depth-first search algorithm, followed by pruning of redundant paths according to equivalency relation introduced in~\cite{bhattachacharya2010homotopyconstraints}, proposed in~\cite{rosmasnn2015multiple} is also affected by this issue. 
For graph search to be efficient, a sparser roadmap, with reduced number of nodes has to be constructed. 
Authors in~\cite{zhang2019sparse} propose a method to delete certain edges from a dense roadmap to construct a sparse near-optimal graph. 
However, even though created graph is sparser, it still contains the same amount of nodes, thus still resulting in high number of redundant paths being found.

Visibility-PRM~\cite{simeon2000visibilityprm} is a variant of PRM that constructs a sparse roadmap, while discarding some of the nodes as well, resulting in a roadmap more efficient and compact compared to traditional PRM. 
It does so by introducing a concept of visibility domains. Every domain is defined by a ``guard" that covers a space ``visible" to the guard. 
No guards are allowed to be visible to each other, thus, they are connected through additional samples called ``connectors". 
The method in~\cite{schmitzberger2002hppr} extends the original Visibility-PRM by allowing creation of cycles but keeping the roadmap simply connected, making the method suitable for distinct path searching. 
However, it may not capture all \changed{topological} classes, especially in environments containing multiple narrow passages. 
In this scenario a connector node has to be sampled exactly inside a narrow passage, but only after two guard nodes have already been created in specific locations.


Authors in~\cite{simeon2008deformation} modify the original Visibility-PRM by iteratively adding a limited number of cycles, capturing new \changed{topological} classes, by connecting components of visible sub-roadmap. 
Even though the method shows promising results, determining a visible sub-roadmap and its separate components gets progressively more computationally expensive in complex environments. 


The algorithm in~\cite{zhou2021raptor} was designed for fast trajectory re-planning, but includes the sub-task of finding \changed{topologically} distinct paths. 
It modifies Visibility-PRM algorithm to make it computationally efficient, by discarding many generated samples.
This method achieves best results in open scenarios, however is very susceptible to initial placement of new guard nodes in scenarios where visibility is limited, which affects both computational speed and functionality.


The method in~\cite{penicka2022quadrotor} tackles minimum-time trajectory planning problem, but also proposes a solution for finding multiple paths with distinct \changed{topological} classes. 
The algorithm starts by iteratively searching for the shortest path in a constructed PRM using Dijkstra's. 
For each path found, a region around the node with smallest clearance is removed from the roadmap. 
This process is repeated until no new path can be found. 
To address cases where multiple distinct paths pass through a deleted region, algorithm is called recursively. 
The limitation of this method is the lack of information required to optimally select a region to remove, failing to find some of the paths as a result. 
Recursion may also lead to combinatorial explosion, drastically increasing run-time.

The main limitation of existing methods is their inconsistency across different environments, leading to significant variations in their performance, especially in challenging environments that contain multiple narrow passages. 
CTopPRM is capable of efficiently reducing a dense roadmap, required to accurately represent such environment, by dividing it into clusters. 
This significantly reduces both number of edges and nodes, while maintaining the same connectivity of free space, as the initial roadmap. 
This allows it to both effectively and consistently identify a large number of \changed{topologically} distinct paths, even in challenging environments.

\section{Problem Statement\label{sec:problem}}

The goal of this paper is to tackle the problem of finding multiple \changed{topologically} distinct paths, i.e., paths between same endpoints going around an obstacle from different sides. 
By identifying multiple such paths, the robot has greater flexibility when planning its movements.
We assume a 3D configuration space $\mathcal{C}$, where each configuration $q=(x,y,z) \in \mathcal{C}$ is the position of the robot. 
Let $\mathcal{C}_{\mathrm{free}} \subseteq  \mathcal{C}$ denote the set of  collision-free configurations.
A path $\pi(s)$ is a continuous curve that connects $q_{start} \in \mathcal{C}_{\mathrm{free}}$ and $q_{goal} \in \mathcal{C}_{\mathrm{free}}$, and is denoted as collision-free if $\pi(s) \in \mathcal{C}_{\mathrm{free}}, \forall s \in [0,1]$.

\changed{A common definition of topological equivalency is homotopy~\cite{hatcher2002topology}, where two continuous functions from one topological space to another are considered equivalent if one can be smoothly deformed into the other.}
This definition, applied on paths, was summarized in~\cite{simeon2008deformation}. 
Homotopy of two paths $\pi(s)$ and $\pi'(s)$ in $\mathcal{C}$ is said to exist if there is a continuous map $h:[0,1]\times[0,1] \xrightarrow{} \mathcal{C}_{\mathrm{free}}$ such that $h(s,0) = \pi(s)$, $h(s,1) = \pi'(s)$ for all $s \in [0,1]$, and $h(0,t) = h(0,0)$ and $h(1,t) = h(1,0)$ for all $t \in [0,1]$.

While homotopy is a widely used concept, it has been found to be inadequate for capturing a sufficient number of useful paths in $\mathbb{R}^3$ space. 
To address this limitation,~\cite{simeon2008deformation} introduced the concept of Visibility Deformation (VD) which captures more useful paths. 
Unlike homotopy, which preserves the topology of the paths, VD focuses on preserving certain visibility-related properties of the path, such as the ability to evade obstacles, effectively reducing dimensionality of deformation between the paths. 
However, the approach is still computationally expensive. 
Therefore,~\cite{zhou2021raptor} proposes an extension to VD called Uniform Visibility Deformation (UVD), which is more efficient.
\begin{definition}
Two paths $\pi(s)$, $\pi'(s)$ parameterized by $s\in[0,1]$ and satisfying $\pi(0) = \pi'(0)$, $\pi(1) = \pi'(1)$, belong to the same uniform visibility deformation class, if for all s, the line-segment from $\pi(s)$ to $\pi'(s)$ is collision-free.
\end{definition}

\changed{In this paper, we tackle the problem of finding a set $\Pi$ of distinct topological paths that belong to different UVD classes, using a combination of motion planning methods, as well as clustering and graph-based search algorithms.}

\section{Clustering Topological PRM\label{sec:method}}

\input{fig/blender}


The proposed method we named Clustering Topological PRM (CTopPRM) finds distinct topological paths using a hierarchical approach that starts by constructing a dense roadmap using Informed-PRM~\cite{aria2021informedprm}. 
The nodes in the roadmap are divided into two initial clusters (that are defined by $q_{start}$ and $q_{goal}$), and more clusters are iteratively identified.
In each iteration, new cluster centroid is created between two neighbouring clusters.
Cluster centroids are then used as vertices of new sparse roadmap, which is then searched for paths with diverse uniform visibility deformation classes. 
Finally, found paths are shortened and filtered. 
The algorithm uses Euclidean Signed Distance Field (ESDF) for collision checking. 
The method is summarized in Algorithm~\ref{alg:alg1} and its visualization is shown in Figure~\ref{fig:blender}. 




\begin{algorithm}[!ht]
\small
\DontPrintSemicolon
  \KwInput{$q_{start}$, $q_{goal}$}
  \KwData{M max clusters, $\kappa_p$ DFS termination condition, $\kappa_s$ pruning parameter}
  \KwOutput{$\Pi$ = ($\pi_1$, $\pi_2$, ..., $\pi_n$) - found topological paths\\\vspace{0.05cm} \hrule\vspace{0.05cm}}
    ($V$, $E$), $l$ $\xleftarrow{}$ \textbf{Informed-PRM}($q_{start}$, $q_{goal}$)\nllabel{alg1:lineprm} \textit{//} \cite{aria2021informedprm}\;  
    $C_V$ $\xleftarrow{}$ \{$q_{start}$, $q_{goal}$\}\;
    \changed{can\_add $\xleftarrow{}$ true\;
    \While{can\_add \textbf{AND} $|C_V|$ $<$ M}{
        ($V$, $E_C$) $\xleftarrow{}$ \textbf{clusterGraph}(($V$, $E$), $C_V$) \textit{// Alg.~\ref{alg:alg2}}\;
        $C_V$, can\_add $\xleftarrow{}$ \textbf{addCentroids}(($V$, $E_C$), $C_V$) \textit{// Alg.~\ref{alg:alg3}}\;
    }
    $C_E$ $\xleftarrow{}$\textbf{findClusterEdges($C_V$, $E_C$)}\nllabel{alg3:clusteredges}\;}
    $\Pi^d$ $\xleftarrow{}$ \textbf{findDistinctPaths}($C_V$, $C_E$, $\kappa_p\cdot l$)\nllabel{alg1:find}\;
    $\Pi$ $\xleftarrow{}$ \textbf{filterPaths}($\Pi^d$)\nllabel{alg1:filter}\;
\caption{CTopPRM}
\label{alg:alg1}
\end{algorithm}


\changed{
Algorithm~\ref{alg:alg1} starts with \textbf{Informed-PRM} constructing a graph ($V$, $E$) with vertices $V$ and edges $E$.
This graph is then divided into clusters by \textbf{clusterGraph} method, transforming the roadmap into a forest ($V$, $E_C$). 
Method \textbf{addCentroids} defines a node as a new cluster centroid.
Roadmap is then clustered again. 
This process is iteratively repeated until a suitable new centroid candidate was not found, or a predefined maximum number of clusters M is reached resulting in a fully clustered roadmap shown in Figure~\ref{fig:blender}(b).
Then, a low-order graph is created with cluster centroids as vertices. 
Edges connecting them are found using the \textbf{findClusterEdges} method, which transforms saved shortest paths between each pair of neighbouring clusters into edges, finalizing the construction of the low-order graph, shown in Figure~\ref{fig:blender}(c).
Method \textbf{findDistinctPaths} searches this graph for a set of distinct paths $\Pi^d$, which are then shortened, \changed{akin to the approach} in~\cite{zhou2021raptor} and~\cite{penicka2022quadrotor}, and filtered by method \textbf{filterPaths}.
\changed{Each part of the CTopPRM algorithm is explained in more detail in the following subsections.}
}

\subsection{Dense Probabilistic Roadmap Construction}\label{subsection:PRM}
The goal of this step of the CTopPRM algorithm is to densely represent free-space $\mathcal{C}_{\mathrm{free}}$ which
is realized using Informed-PRM~\cite{aria2021informedprm}.
Each vertex is connected to its k-nearest neighbours using a straight-line, if possible.
The shortest path in the constructed roadmap is then found using Dijkstra; let $l$ denote its length.

\subsection{Graph clustering} \label{subsection:clustering}

The \textbf{clusterGraph} method described in Algorithm~\ref{alg:alg2} divides roadmap into clusters with shortest-path tree~\cite{dijkstra1959note} structure, with each cluster centroid being root of each shortest-path tree.
All clusters together form a shortest-path forest~\cite{dial1969SPforest}.
In a shortest-path forest, each tree represents the shortest paths from all nodes to their closest root.
\begin{algorithm}[!ht]
\small
\DontPrintSemicolon
  \KwInput{($V$, $E$) roadmap, $C_V$ cluster centroids}
  \KwOutput{($V$, $E_C$) clustered roadmap\\\vspace{0.05cm} \hrule\vspace{0.05cm}} 
    \KwFor{v $\in$ V}{
        v.value $\xleftarrow{} \infty$, v.cluster $\xleftarrow{}$ -1\;
    }
    \KwFor{c $\in C_V$}{
        c.value $\xleftarrow{}$ 0, c.cluster $\xleftarrow{}$ cluster\_id, cluster\_id++\;
    }
    heap $\xleftarrow{} V\cup C_V$\;
    \While{heap $\neq \emptyset$}{
        v $\xleftarrow{}$ heap.pop()\;
        \KwFor{neighbour n of node v}{
            value$_{new} \xleftarrow{}$ v.value + cost(v, n)\nllabel{alg2:begin}\;
            \If{value$_{new} <$ n.value}{
                n.value $\xleftarrow{}$ value$_{new}$\;
                n.cluster $\xleftarrow{}$ v.cluster, n.parent $\xleftarrow{}$ v\nllabel{alg2:end}\;
            }
            \ElseIf{v.cluster $\neq$ n.cluster}{
                cost$_{new} \xleftarrow{}$ v.value + cost(v, n) + n.value\nllabel{alg2:cost}\;
                update $\pmin$ and $\pmax$ if necessary\; 
            }
        }
    }
\caption{Graph clustering}
\label{alg:alg2}
\end{algorithm}

The division of the roadmap into clusters is implemented using a min-heap, making the time complexity $\mathcal{O}(E \log V)$. 
The initial cluster centroids are $q_{start}$ and $q_{goal}$. 
Each cluster is expanded from its centroid, creating connections to minimize the total cost of a path from each vertex to the nearest cluster centroid, summarized by Lines~\ref{alg2:begin}-\ref{alg2:end} of Algorithm~\ref{alg:alg2}. 
If two neighbouring vertices belong to different clusters, total cost of path connecting two cluster centroids over these two vertices is calculated, as shown in Line~\ref{alg2:cost}. 
For each pair of neighbouring clusters i and j, the method maintains paths $\pmin$ and $\pmax$ that represent the shortest and the longest paths connecting the two clusters, respectively, with a prospect they might represent different UVD classes. 
Edges that do not belong to either cluster, but are a part of these paths are called minimum and maximum cluster connection, and they are crucial for selecting new cluster centroids in the steps to follow. 
An example of these connections is shown in Figure~\ref{fig:blender}(a), where green line represents minimum and red line represents maximum cluster connection.

\subsection{Adding new centroids} \label{subsection:centroid}

The motivation behind division of PRM into multiple clusters is to create an easily searchable graph with cluster centroids as vertices which will have significantly less vertices than the dense roadmap.
To capture all UVD classes, while minimizing order of the graph, vertices have to be placed methodically. 
CTopPRM's approach to create new cluster centroids is depicted in Algorithm~\ref{alg:alg3}.

It starts by comparing connections $\pmin$ and $\pmax$ for each pair of connected clusters. 
The method \textbf{Deformable} in Line~\ref{alg3:deformable} of Algorithm~\ref{alg:alg3} then checks if these two paths belong to the same UVD class.
Each path is discretized to $n = \lceil \pmax.length / \Delta d \rceil$ steps.
Each line-segment between the points $\pmax[k]$ and $\pmin[k]$, $k=0,\ldots, n$, is tested for collisions with the resolution $\Delta d$.
If they are not deformable, creating a new centroid at the border of these two clusters is beneficial in capturing more UVD classes, as there are two distinct paths connecting two existing cluster centroids. 
The ratio of their lengths is then calculated and saved. 
The connection with the highest ratio is selected and one of the two neighbouring nodes belonging to its maximum connection \changed{(endpoints of the red line in Fig.~\ref{fig:blender}(a))}, is determined as a new centroid \changed{(light yellow in Fig.~\ref{fig:blender}(a))}.
\changed{By adding a new centroid at the maximum connection, a new topologically unique path between given cluster centroids is created, allowing detour through the newly defined centroid, usually resulting in at least one new path between start and goal nodes.}
\changed{If $\pmin$ and $\pmax$ are deformable into each other for all neighbouring clusters, thus a suitable new candidate for the centroid cannot be identified, then the \textit{can\_add} variable remains set to \textit{false}, indicating termination of iteration in Alg.~\ref{alg:alg1}.}

\begin{algorithm}[!ht]
\small
\DontPrintSemicolon
  \KwInput{($V$, $E_C$) clustered roadmap, $C_V$ cluster centroids}
  \KwData{$\Delta d$ collision-detection resolution}
  \KwOutput{$C_V$ set of cluster centroids,  can\_add end criterion\\\vspace{0.05cm} \hrule\vspace{0.05cm}} 
    can\_add $\xleftarrow{}$ false\;
    ratio $\xleftarrow{} \emptyset$ \;
    \KwFor{(i, j) $\in$ connected\_clusters}{
        \If{\textbf{\textup{not Deformable}}($\pmin$, $\pmax$, $\Delta d$)}{ \nllabel{alg3:deformable}
            ratio $\xleftarrow{}$ ratio $\cup \{\frac{\pmax.length}{\pmin.length} \}$ \;
            can\_add $\xleftarrow{}$ true\;}
    }
    new\_centroid $\xleftarrow{}$ \textbf{getCentroid}(\textbf{max}(ratio))\nllabel{alg3:ratio}\;
    $C_V \xleftarrow{} C_V \cup $ new\_centroid\;
\caption{Adding new centroids}
\label{alg:alg3}
\end{algorithm}

\subsection{Path finding and filtering} \label{subsection:filtering}

The method \textbf{findDistinctPaths} in Line~\ref{alg1:find} of Algorithm~\ref{alg:alg1} searches the graph using Depth-first search (DFS) algorithm with visited list similar to~\cite{rosman2017dfs}. 
In this depth-limited depth-first search, the expansion on the actual node is terminated if the current path length is greater than $\kappa_{p}$ times the length of best solution $l$.

To accommodate future applications, e.g., for planning high-speed UAV trajectories along the paths, CTopPRM uses a series of methods, included in \textbf{filterPaths} function in Line~\ref{alg1:filter} of Algorithm~\ref{alg:alg1}, to augment and filter found paths. 
Method similar to shortening in~\cite{zhou2021raptor} and~\cite{penicka2022quadrotor} is used first to shorten found paths in forward and backward pass. 
Any shortened paths longer than a threshold defined as length of the shortest found path multiplied by the parameter $\kappa_{s}$ are then pruned away. 
Doing this filters out any paths that include sub-optimal movement, e.g. looping around obstacles. 
Finally, one last UVD equivalency check is performed on shortened paths to filter out any paths belonging to the same UVD class. 
The final output of the CTopPRM (Algorithm~\ref{alg:alg1}) is a set of paths representing different UVD classes.

\subsection{Discussion} \label{subsection:filtering}

\changed{The CTopPRM uses the probabilistically complete Informed-PRM~\cite{aria2021informedprm} that, for an infinite number of samples, ensures finding a path between start and goal if it exists.}
\changed{Thus, it can capture for an infinite number of samples all distinct topological paths.}
\changed{At the same time, if the number of cluster centroids grows, the probability of finding all distinct paths approaches certainty, as eventually all PRM samples would be searched by the DFS. }
\changed{Yet, for practical purposes, a more efficient approach is favored in the CTopPRM over probabilistic completeness due to application constraints.}

\section{Results\label{sec:results}}

\input{fig/scenarios}

The performance of the CTopPRM is evaluated in three different environments shown in Figure~\ref{fig:scenarios}. 
The purpose of having multiple thematically different environments is to evaluate robustness of our and related methods. 
The most important evaluation metric is the number of UVD classes each method is able to find in form of a path within such a UVD class. 
We also consider computational time and quality of found paths, represented by their respective lengths.

\input{tab/parameters}

CTopPRM is implemented in C++, and experiments are run on AMD Ryzen 7 6800HS CPU. 
Values of parameters used in the test is shown in Table~\ref{tab:parameters}. 

The computational time of all methods examined in this study is primarily influenced by \changed{three} key parameters: $\Delta d$ --- the resolution of collision detection, the number of samples used to build the dense roadmap \changed{and k --- the number of neighbours each node is connected to during the construction of the roadmap}.
\changed{In all conducted experiments, k was set to 14, a value determined experimentally to achieve a balance between computational time and adequate coverage of the free space.}
The size of the spherical robot is specified by the clearance parameter. 
CTopPRM algorithm includes two more parameters: M (maximum number of clusters) and $\kappa_{p}$ (DFS termination condition). 
Both parameters affect the trade-off in performance of the method, reducing computational time, but at a cost of quality and quantity of found paths.

We evaluate the performance of CTopPRM by comparing it with three other methods that tackle the same challenge. 
These include the method we call RAPTOR from~\cite{zhou2021raptor}, which solves the sub-task of identifying distinct topological paths, the Distinct Path Search algorithm proposed in~\cite{penicka2022quadrotor}, referred to as B. spheres, and an approach based on~\cite{simeon2008deformation} called P-D-PRM, adapted slightly to ensure reasonable run-time.

\subsection{Windows environment} \label{subsection:windows}

\input{tab/wind_table}

The first set of experiments is performed in environments called windows which contain a small number of narrow passages (windows) placed on one to three parallel walls, making maximum number of distinct UVD classes, ground truth (GT), easily determinable. 
Name of the maps in Table~\ref{tab:windows} indicates number of windows on each wall.
For example, scenario 1-3-1 shown in Figure~\ref{fig:windows} contains one window on the first wall, three on the second and one on the third. 
Zero indicates a given wall is missing and 's' for 'side' indicates that a specific window is not in the middle of the wall.
These scenarios are tested in $\mathbb{R}^2$ space with a circular robot.
Each algorithm is evaluated on 100 runs in every scenario. 

Figure~\ref{fig:graphs} presents the performance comparison of all algorithms in the 1-3-1 scenario. 
Our method outperformed other methods by finding all three distinct paths more reliably in this particular scenario. 
We report computational time and success rate of each algorithm in finding every single distinct path that exists in a given scenario. 

The results of this experiment are shown in Table~\ref{tab:windows}. 
They indicate that CTopPRM algorithm manages to find all but one path, across all testing scenarios, with highest success rate. 
Additionally, CTopPRM finds most paths with a success rate close to \SI{100}{\percent}, with lowest success rate being \SI{70}{\percent}, proving its effectiveness and reliability, both absolutely and relatively to other methods.

\input{fig/graphs}

All the methods performed better in simpler scenarios where only one window is required to be passed (0-2-0 and 0-3-0), and their performance deteriorates as the number of narrow passages increases. 
Visibility-based methods P-D-PRM and RAPTOR demonstrate the most significant decline in performance, with success rates dropping below 10\% for multiple paths in different scenarios.

Interesting results arise from scenario 1-3-0 where P-D-PRM and RAPTOR find the shortest path $\pi_1$ in every run, due to $q_{goal}$ being visible from $q_{start}$, but fail to ever identify any of the remaining paths $\pi_2$ and $\pi_3$.  
Due to the layout of the map, we can assume that Visibility-PRM blocks itself off by placing a guard node in an unfavorable position. 
Additionally, B.spheres method, which otherwise shows more competitive results, is also able to detect these paths with less than \SI{10}{\percent} success rate. 
Contrarily, CTopPRM detects both paths in \SI{70}{\percent} of runs.
\changed{While this result significantly surpasses other methods, it is the most challenging map for the CTopPRM.}
\changed{This is primarily due to the CTopPRM's limitation in capturing UVD classes not included in the initial PRM, which depends on the number of random samples.}
\changed{This weakness is magnified in this test case where closely spaced walls and large free space result in fewer nodes being generated in critical areas.
Moreover, for the tests in Table~\ref{tab:windows}, the algorithm was always terminated  only if all connections between clusters were deformable.
Therefore, Table~\ref{tab:windows} accurately represents the probability of CTopPRM failing to capture a UVD class.
Despite this limitation, CTopPRM clearly outperforms other methods, showcasing its robustness in challenging scenarios.}

The run-time of all methods depends not only on the size of the input roadmap, but also on number of paths \changed{identified}, as most of computational time is taken by filtering process described in Section~\ref{subsection:filtering}. 
This is why scenario 0-3-0 is interesting to us, since all tested methods identify similar amount of paths. 

\input{fig/samples}

As already mentioned, the performance of the methods depends on the number of random samples used to create the initial roadmap. 
We show the performance with the increasing number of random samples in Figure~\ref{fig:samples}. 
They indicate that with the lower number of samples, the methods B. spheres and CTopPRM manage to find more paths, but are clearly slower than both P-D-PRM and RAPTOR. 
Additionally, it is important to note that P-D-PRM finds significantly fewer paths than the other methods, because it consists of two phases, which have to share the total amount of samples. 
With a growing number of samples, the performance of all algorithms in terms of the number of paths found converges towards three, which is the ground truth in this scenario. 
However, unlike B. spheres and CTopPRM, which record just a minor increase in computational time, both P-D-PRM and RAPTOR become significantly slower.

Overall, CTopPRM shows computational speed competitive with other related methods, while clearly outclassing them in terms of path detection success rate. 
Therefore, CTopPRM proves to be the most efficient and effective in scenarios with smaller number of distinct UVD classes.

\subsection{Complex environments} \label{subsection:complex}

\input{tab/all_maps}

The second set of experiments was conducted in complex environments, containing a high number of distinct UVD classes. 
The first, called ``poles" and depicted in Figure~\ref{fig:poles}, resembles a small forest-like area, while the second, shown in Figure~\ref{fig:building} and called ``building", requires a robot to traverse a closed, multi-level building area through doors and windows. 
For each of these environments, we tested the performance of the methods in three different scenarios with different start and goal configurations ($q_{start}$ and $q_{goal}$).
Each method was tested in 100 runs in each scenario.
The poles scenarios are tested in $\mathbb{R}^2$ space with a circular robot and building scenarios are tested in $\mathbb{R}^3$ space with a spherical robot.

Table~\ref{tab:all} summarizes performance of the methods in terms of computational time, the quantity of found paths represented by the highest number of paths found in a single run (best) and average number of found paths across all 100 runs. 
The quality of paths is evaluated as an average length of n-shortest paths found over all 100 runs, where $n$ is indicated in the table. 
This metric is supposed to show if a method is able to consistently find $k$ shortest paths in each scenario.

The results show that CTopPRM performs the best in terms of quantity of paths found, identifying the most paths in a single run in every scenario, as well as significantly outscoring other methods in average number of paths found in all scenarios. 
Additionally, CTopPRM also outperforms other methods in terms of quality of found paths in all but one scenario in poles environment, where it records a score just 1\% worse than RAPTOR. 
CTopPRM is also the only method to find at least $n$ paths across 100 runs in every single scenario.
Failure to do so is denoted by N/A in Table~\ref{tab:all}.



P-D-PRM and RAPTOR achieve results comparable to CTopPRM in poles environment.
Yet, they fail to identify a single path in majority of runs in building environment.
P-D-PRM is even unable to detect a single path across all 100 runs for a whole scenario. 
This is caused by the increased visibility in poles environment, allowing long connections for Visibility-PRM based methods. 
Contrarily, building environment consists of multiple narrow passages, which was shown in Section~\ref{subsection:windows} to be unfavourable.
CTopPRM delivers consistent results in all environments, verifying its robustness.

\changed{The distribution of CTopPRM's computational time was studied in the poles environment. 
Experimental results indicate that Informed-PRM's construction takes \SI{6.31}{\ms}, clustering lasts \SI{1.72}{\ms}, path search is \SI{0.09}{\ms}, and path filtering \SI{6.23}{\ms}. 
Path searching and filtering process exhibit a significant standard deviation (\SI{0.09}{\ms} and \SI{2.62}{\ms}, respectively), influenced by the quantity of discovered paths.}

In overall, CTopPRM algorithm outperformed other methods in vast majority of scenarios in computational time, quality and quantity of found paths.
It also has the best trade-off between computational time and number of paths found.

\section{Conclusions\label{sec:conclusion}}

This paper introduced a new sampling-based method named CTopPRM for finding multiple paths with distinct UVD classes in cluttered environments. 
The CTopPRM clusters initially sampled dense roadmap in order to efficiently simplify the search of multiple distinct paths.
Through testing in a variety of environments, we demonstrated that CTopPRM is both efficient and robust. 
In majority of test cases, it surpassed other related methods in number of found distinct paths, and their length, during similar computational time.
We improved the average number of \changed{topological} classes detected within the same run-time by \SI{30}-\SI{300}{\percent}, depending on the scenario. 
Additionally, the CTopPRM allows controlling the balance between computational time, quantity and quality of found paths, making it highly adaptable for online planning.
As future work, we aim to extend CTopPRM to enable fast trajectory re-planning, and to deploy it online on unmanned aerial vehicles (UAVs) to test high-speed flight in partially unknown environments.

\balance

\bibliographystyle{IEEEtran}
\bibliography{main}

\end{document}

%% file: fig/blender.tex
\begin{figure*}[htbp]
    \vspace{0.5em}
     \centering
     \footnotesize
     \setlength\tabcolsep{0.2pt}
      \begin{tabular}{ccc}
      \includegraphics[width=0.33\textwidth]{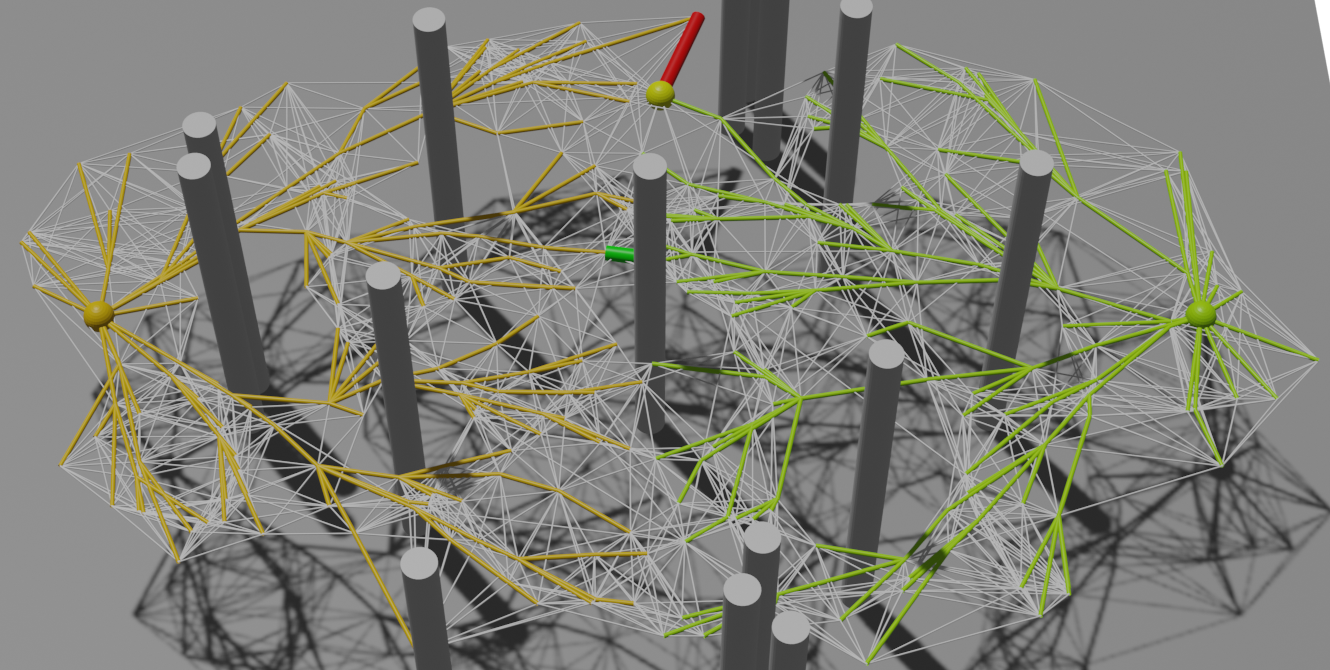} &
      \includegraphics[width=0.33\textwidth]{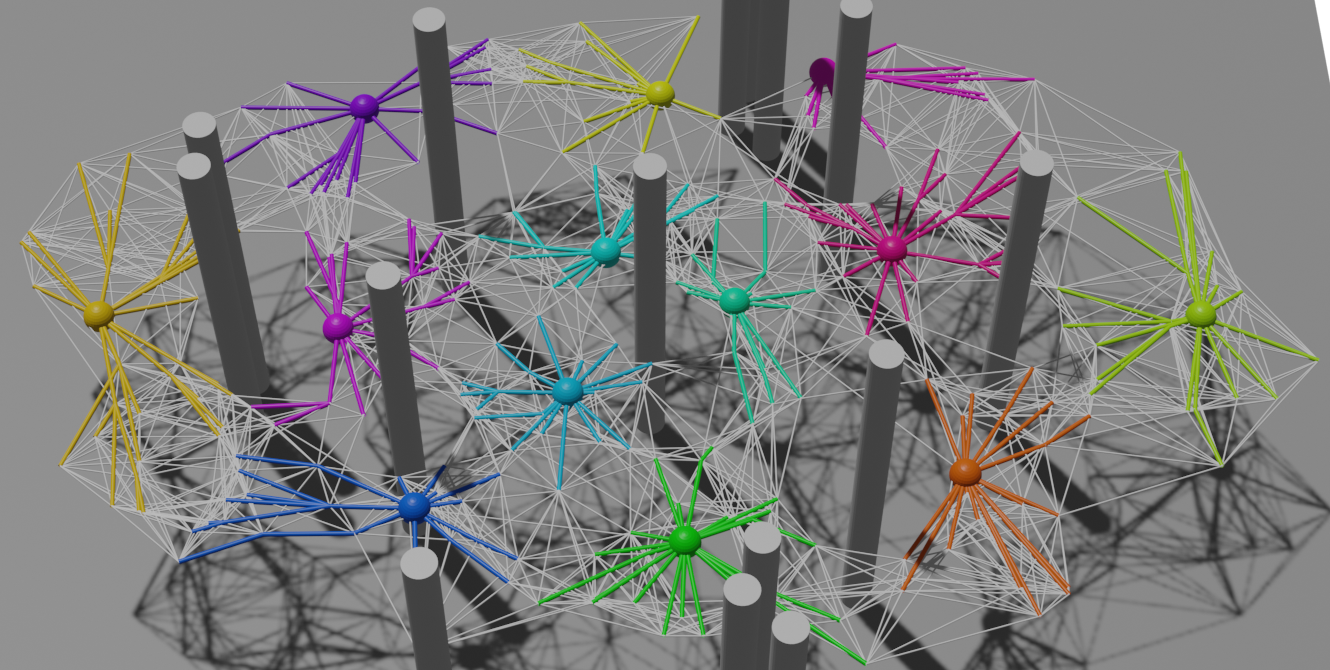} &
      \includegraphics[width=0.33\textwidth]{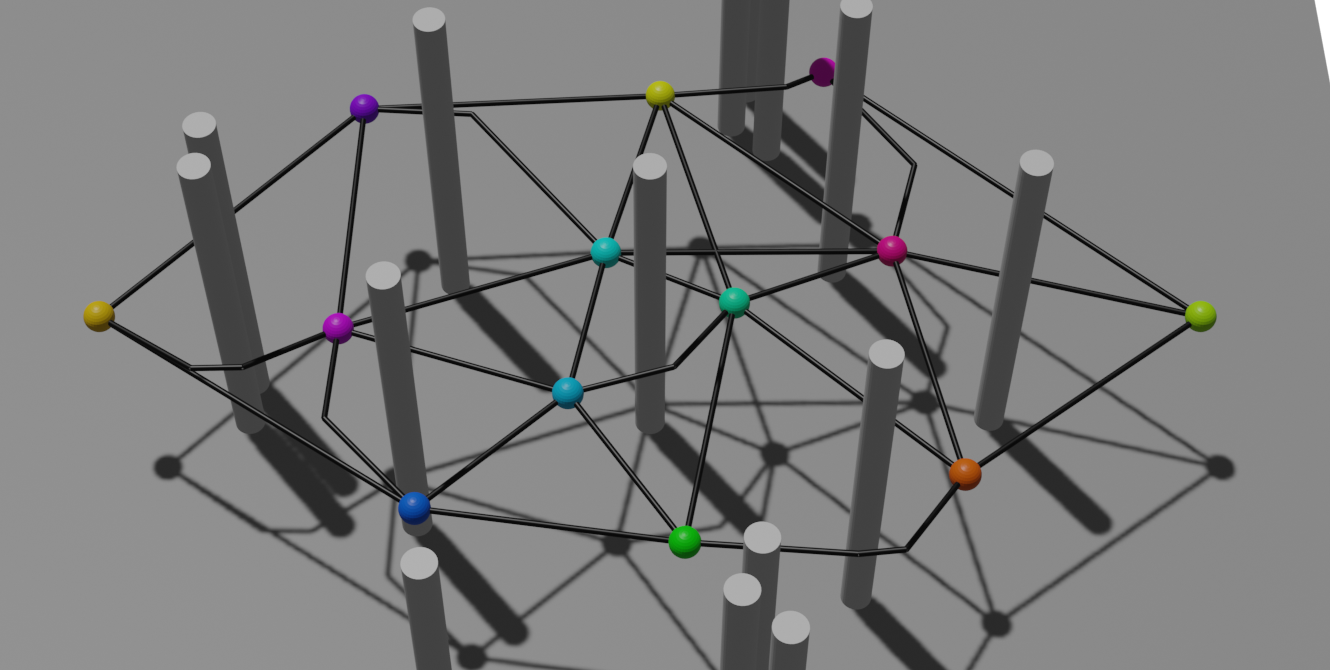} \\[-0.3em]
      (a) roadmap clustered with initial centroids &
      (b) fully clustered roadmap &
      (c) connections between clusters \\[-0.05em]
      \includegraphics[width=0.33\textwidth]{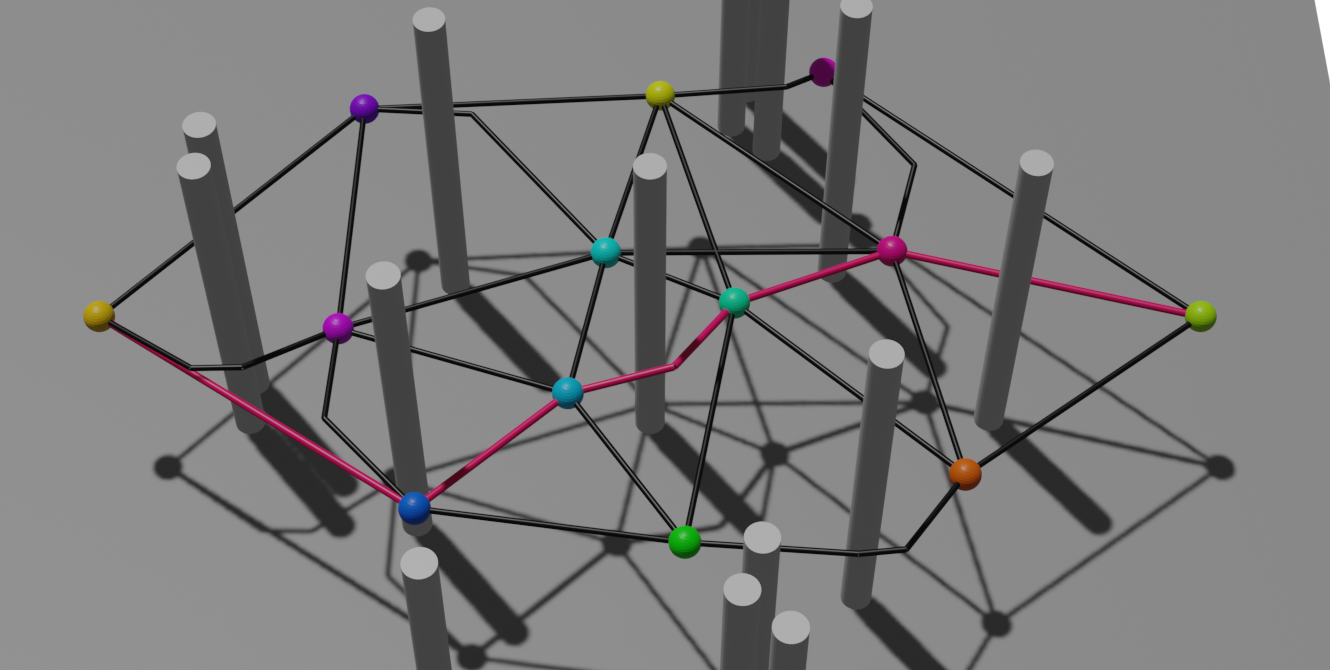} &
      \includegraphics[width=0.33\textwidth]{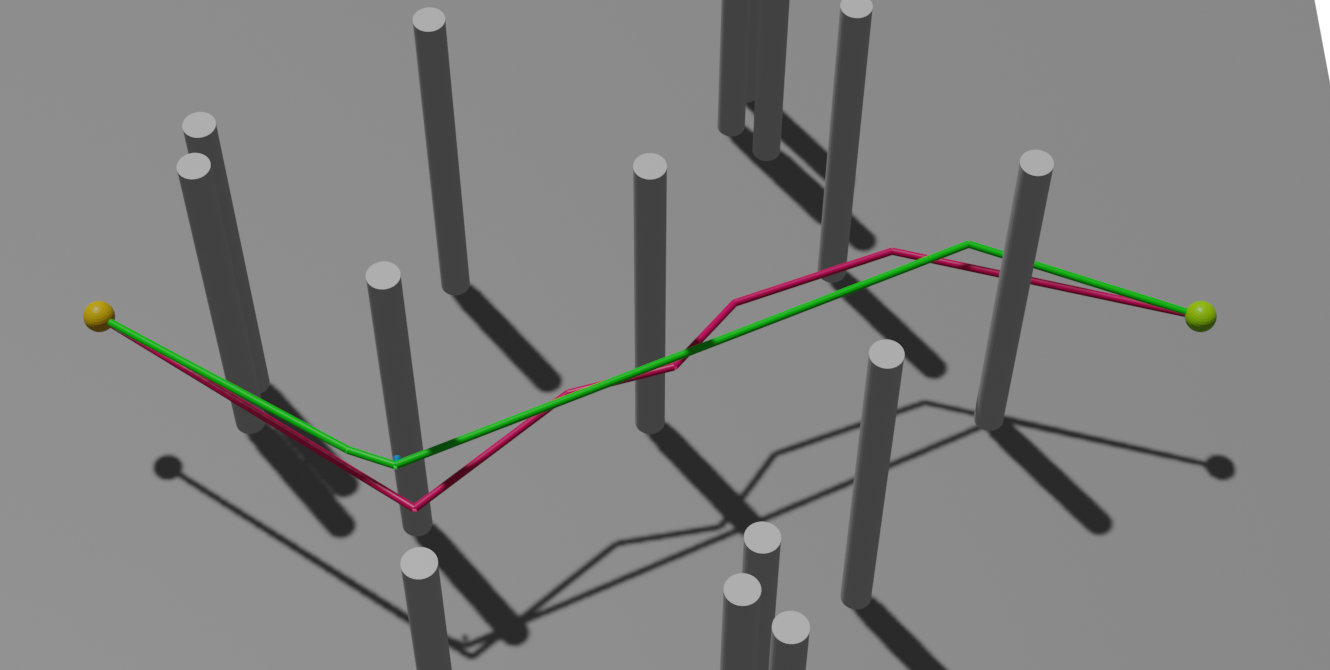} &
      \includegraphics[width=0.33\textwidth]{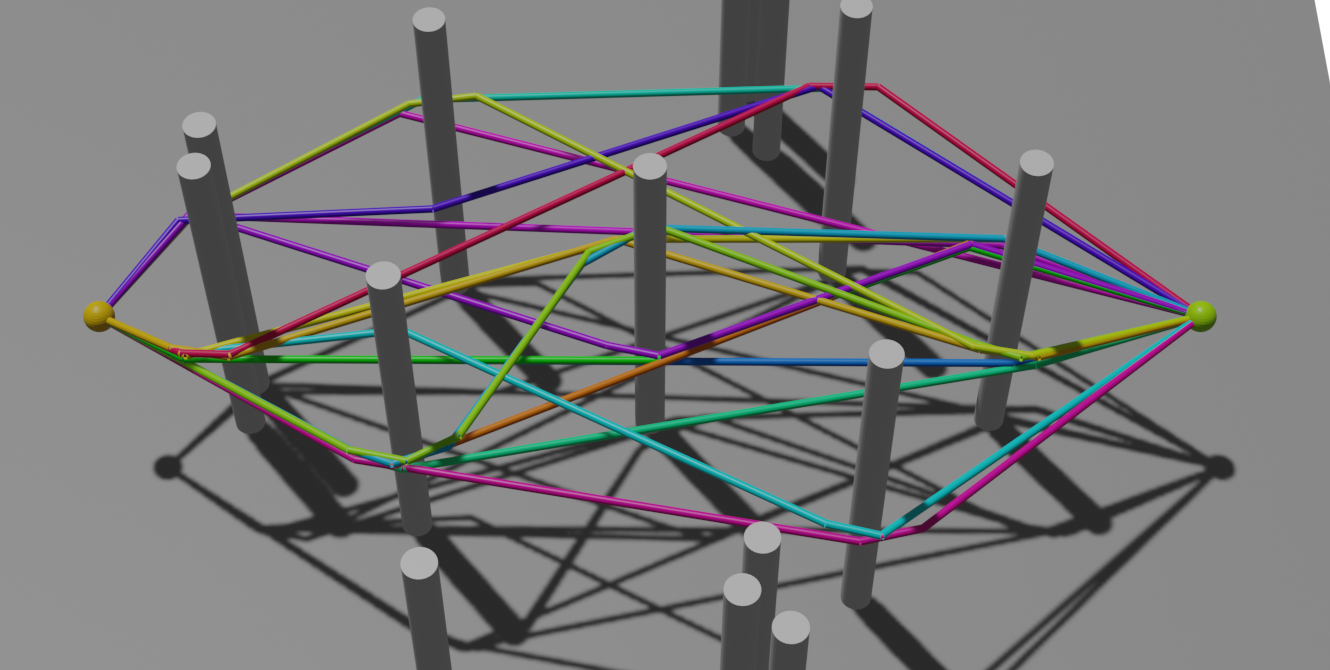} \\[-0.3em]
      (d) one of the found paths &
      (e) shortening of the found path &
      (f) filtered distinct paths
        \end{tabular}
        \vspace{-0.5em}
        \caption{Visualization of individual stages of the algorithm. Generated PRM is first divided into two clusters with $q_{start}$ and $q_{goal}$ as centroids (a). Minimum and  maximum connections, marked by green and red line, are compared and new centroids \changed{(light yellow in (a))} are iteratively created at one of the maximum connections until roadmap is fully clustered (b). Connections between cluster centroids are found and a low-order, sparse graph (c) is constructed and then searched for distinct paths (d). \changed{Each found path is then shortened (e), where red path is shortened into the green. Finally, redundant path are filtered out (f).}}
        \label{fig:blender}
        \vspace{-2.0em}
\end{figure*}

%% file: fig/scenarios.tex
\begin{figure*}[ht]
     \centering
     \begin{subfigure}[b]{0.353\textwidth}
        \centering
        \begin{tikzpicture}
            \node (img) {\includegraphics[width=\textwidth]{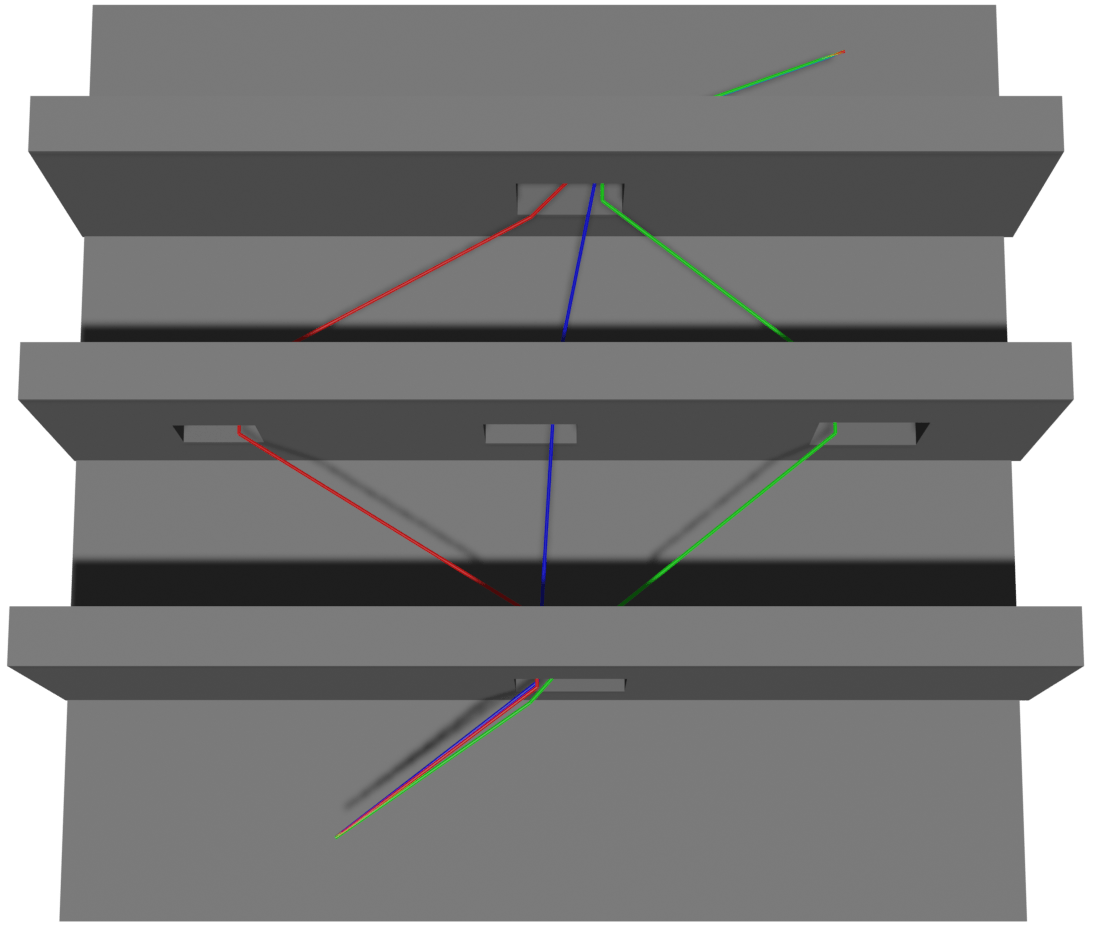}};
            \draw (0.2,-0.2) node {\large $\pi_1$};
            \draw (1.3, 1) node {\large $\pi_2$};
            \draw (-1.44, 1) node {\large $\pi_3$};
         \end{tikzpicture}
         \caption{windows 1-3-1 scenario (from \cite{vonasek2019guidingpaths})}
         \label{fig:windows}
     \end{subfigure}
     \hfill
     \begin{subfigure}[b]{0.3\textwidth}
         \centering
         \includegraphics[width=\textwidth]{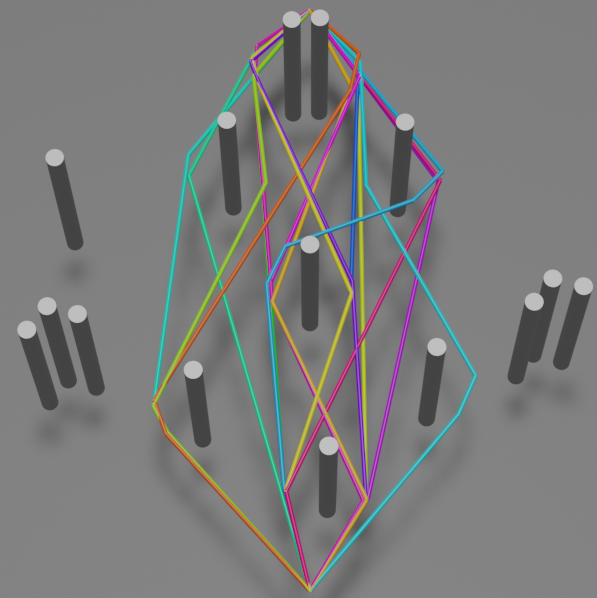}
         \caption{poles}
         \label{fig:poles}
     \end{subfigure}
     \hfill
     \begin{subfigure}[b]{0.295\textwidth}
         \centering
         \includegraphics[width=\textwidth]{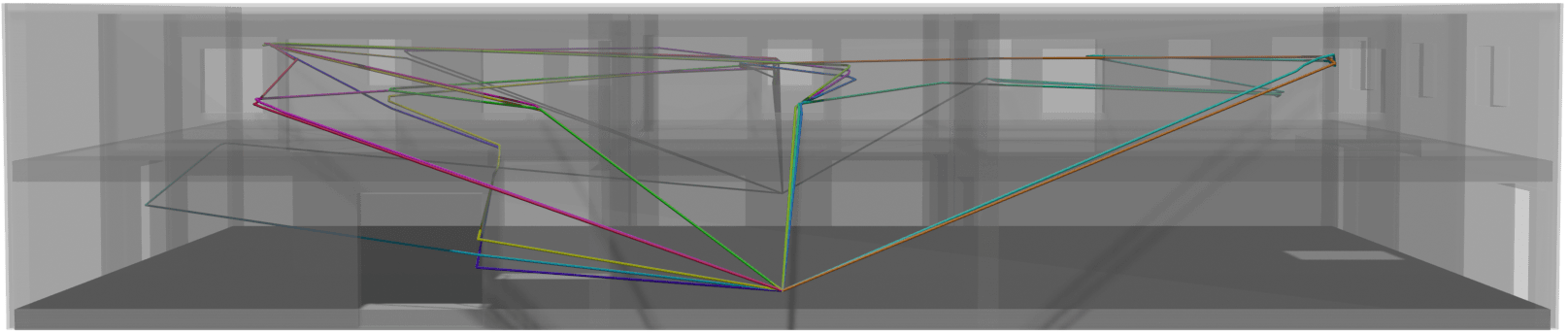}
         \includegraphics[width=\textwidth]{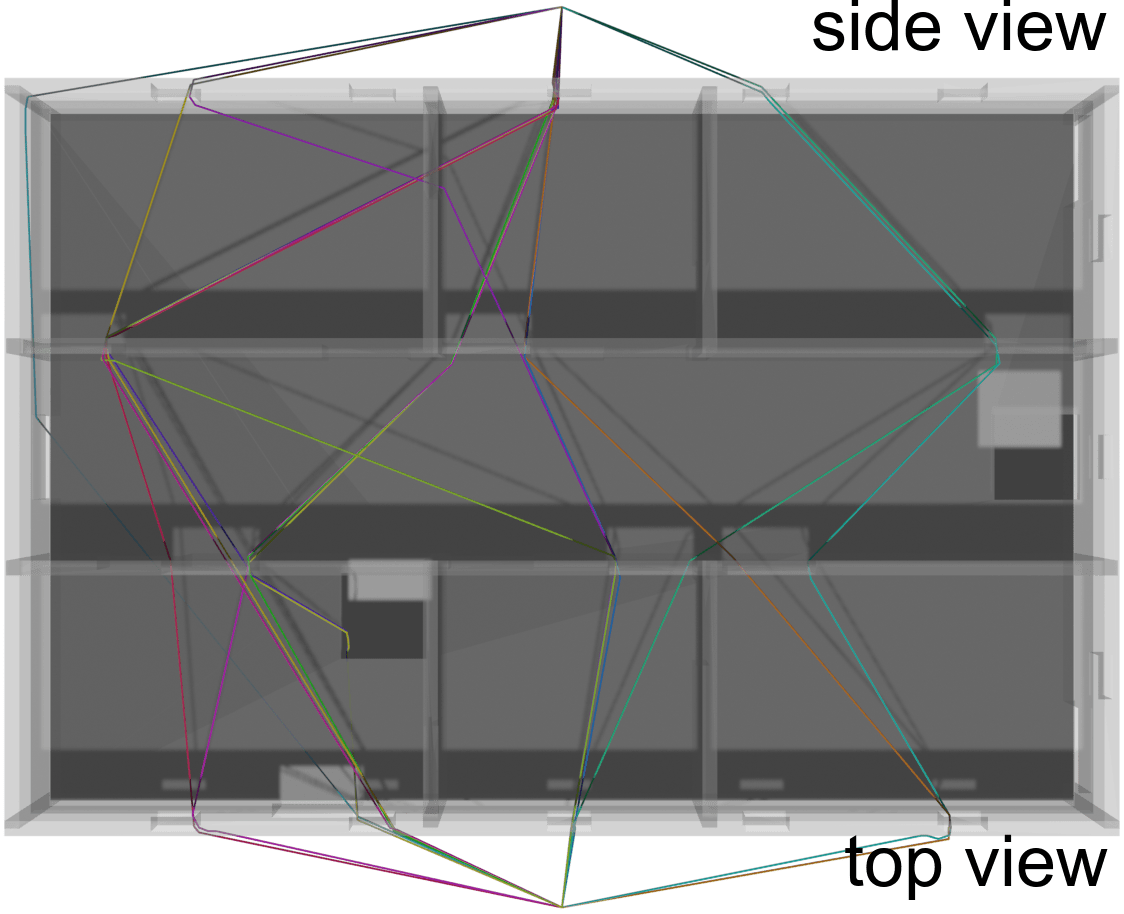}
         \caption{building}
         \label{fig:building}
     \end{subfigure} 
      \vspace{-0.5em}
        \caption{Maps of the environments used for evaluating CTopPRM and related algorithms with paths found by CTopPRM.
        \vspace{-1.8em}}
        \label{fig:scenarios}
\end{figure*}
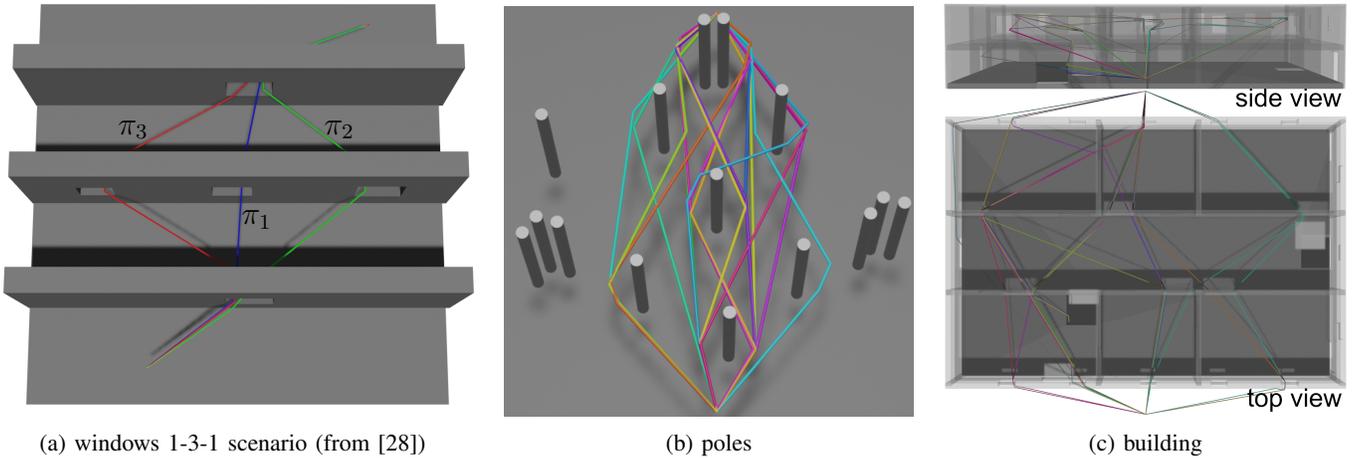

%% file: tab/parameters.tex
\begin{table}[!htb] 
    \centering
    \footnotesize 
    {\renewcommand{\tabcolsep}{5.8pt} 
    \caption{Algorithm parameters \& map size\label{tab:parameters}} 
    \vspace{-1em} 
    \begin{tabular}{crccc}
    \toprule 
    & \textbf{} &  windows & poles & building\\
    & map size & 27x26.7x8 & 10x10x2.8 & 30x20x6.3 \\\midrule
    \multirow{4}{*}{All methods} & clearance &  0.3 & 0.3 & 0.2\\
     & $\Delta d$ & 0.1 & 0.2 & 0.2\\
    & PRM samples & 500 & 300 & 1000\\
    & $\kappa_{s}$ & 1.5 & 1.2 & 1.5 \\\midrule
    \multirow{3}{*}{CTopPRM} & M & 9 & 20 & 20 \\
    & $\kappa_{p}$ & 1.8 & 1.6 & 1.8 \\
    & \changed{k} & \changed{14} & \changed{14} & \changed{14} \\
    \bottomrule
    \end{tabular} 
    } 
    \vspace{-1.0em}
\end{table} 

%% file: tab/wind_table.tex
\begin{table*}[!htb] 
    \vspace{0.5em}
   \centering 
   \footnotesize 
   {\renewcommand{\tabcolsep}{2.85pt} 
   \caption{Success rate of finding each distinct path $\pi_i \in \Pi_{\mathrm{env}}$ in windows environments\label{tab:windows}} 
   \vspace{-0.8em} 
   \begin{tabular}{r|c|c|ccc|c|ccc|c|ccc|c|ccc}
     \toprule 
    &  & \multicolumn4{c|}{\textbf{CTopPRM}} & \multicolumn4{c|}{RAPTOR\cite{zhou2021raptor}} & \multicolumn4{c|}{B. spheres\cite{penicka2022quadrotor}} & \multicolumn4{c}{P-D-PRM\cite{simeon2008deformation}} \\ 
    \midrule
    env. & \textbf{GT} & c.t.[\si{ms}] & $\pi_1$[\si{\percent}] & $\pi_2$[\si{\percent}] & $\pi_3$[\si{\percent}] & c.t.[\si{ms}] & $\pi_1$[\si{\percent}] & $\pi_2$[\si{\percent}] & $\pi_3$[\si{\percent}] & c.t.[\si{ms}] & $\pi_1$[\si{\percent}] & $\pi_2$[\si{\percent}] & $\pi_3$[\si{\percent}] & c.t.[\si{ms}] & $\pi_1$[\si{\percent}] & $\pi_2$[\si{\percent}] & $\pi_3$[\si{\percent}] \\ 
       \midrule
0-2-0 &2 &41 &\textbf{100} &\textbf{100} &- &75 &80 &64 &- &\textbf{36} &99 &94 &- &39 &54 &46 &- \\ 
1-2-0 &2 &54 &\textbf{99} &\textbf{100} &- &65 &51 &42 &- &55 &89 &94 &- &\textbf{52} &35 &24 &- \\ 
1-2-1 &2 &51 &\textbf{99} &\textbf{99} &- &\textbf{35} &4 &7 &- &51 &95 &58 &- &43 &9 &8 &- \\ 
1s-2-1s &2 &48 &\textbf{98} &\textbf{97} &- &42 &40 &26 &- &86 &87 &91 &- &\textbf{40} &42 &9 &- \\ 
\midrule
0-3-0 &3 &70 &94 &\textbf{100} &\textbf{96} &148 &94 &\textbf{100} &92 &53 &\textbf{98} &98 &93 &\textbf{49} &86 &98 &81 \\ 
1-3-0 &3 &51 &\textbf{100} &\textbf{71} &\textbf{70} &101 &\textbf{100} &0 &0 &41 &95 &2 &3 &\textbf{38} &\textbf{100} &0 &0 \\ 
1-3-1 &3 &47 &\textbf{100} &\textbf{89} &\textbf{89} &43 &41 &12 &10 &51 &90 &78 &52 &\textbf{41} &41 &9 &5 \\ 
1s-3-1s &3 &54 &\textbf{100} &\textbf{99} &\textbf{81} &46 &86 &16 &14 &85 &88 &89 &68 &\textbf{21} &89 &9 &7 \\ 
\bottomrule
   \end{tabular} 
   } 
   \vspace{-2em}
\end{table*} 

%% file: fig/graphs.tex
\begin{figure}[!htb]
    \vspace{-0.5em}
    \begin{tikzpicture}
        \node (img) {\includegraphics[width=0.97\columnwidth]{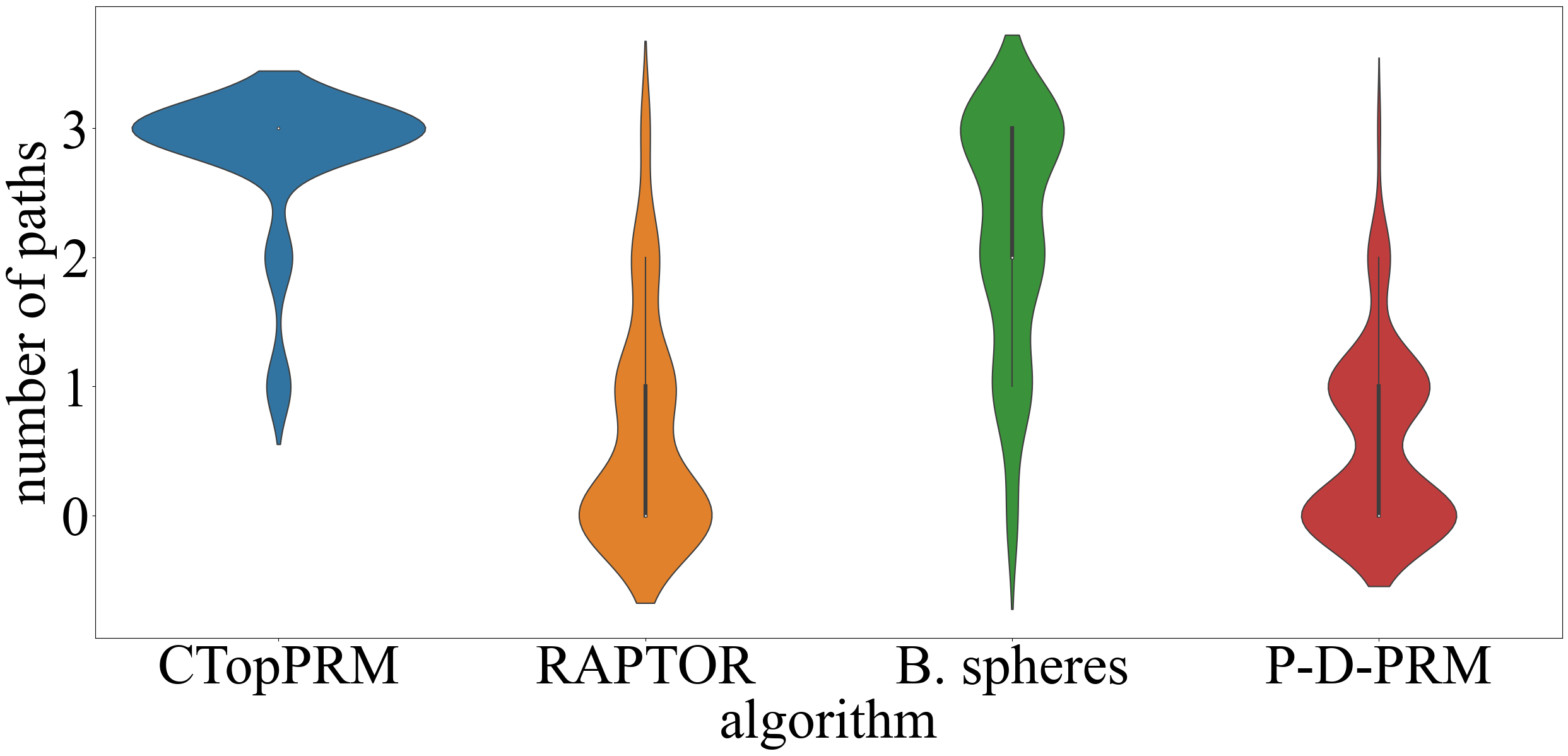}};
        \draw (3.8,1.55) node[black] {\textbf{GT}};
        \draw[black] (-3.7,1.37)  -- (4.17,1.37);
    \end{tikzpicture}
    \caption{Number of paths found in 1-3-1 scenario with highlighted ground-truth (GT) (Fig.~\ref{fig:windows}).}
     \vspace{-1.3em}
     \label{fig:graphs}
\end{figure}

%% file: fig/samples.tex
\begin{figure}[!htb]
     \centering
     \includegraphics[width=\columnwidth]{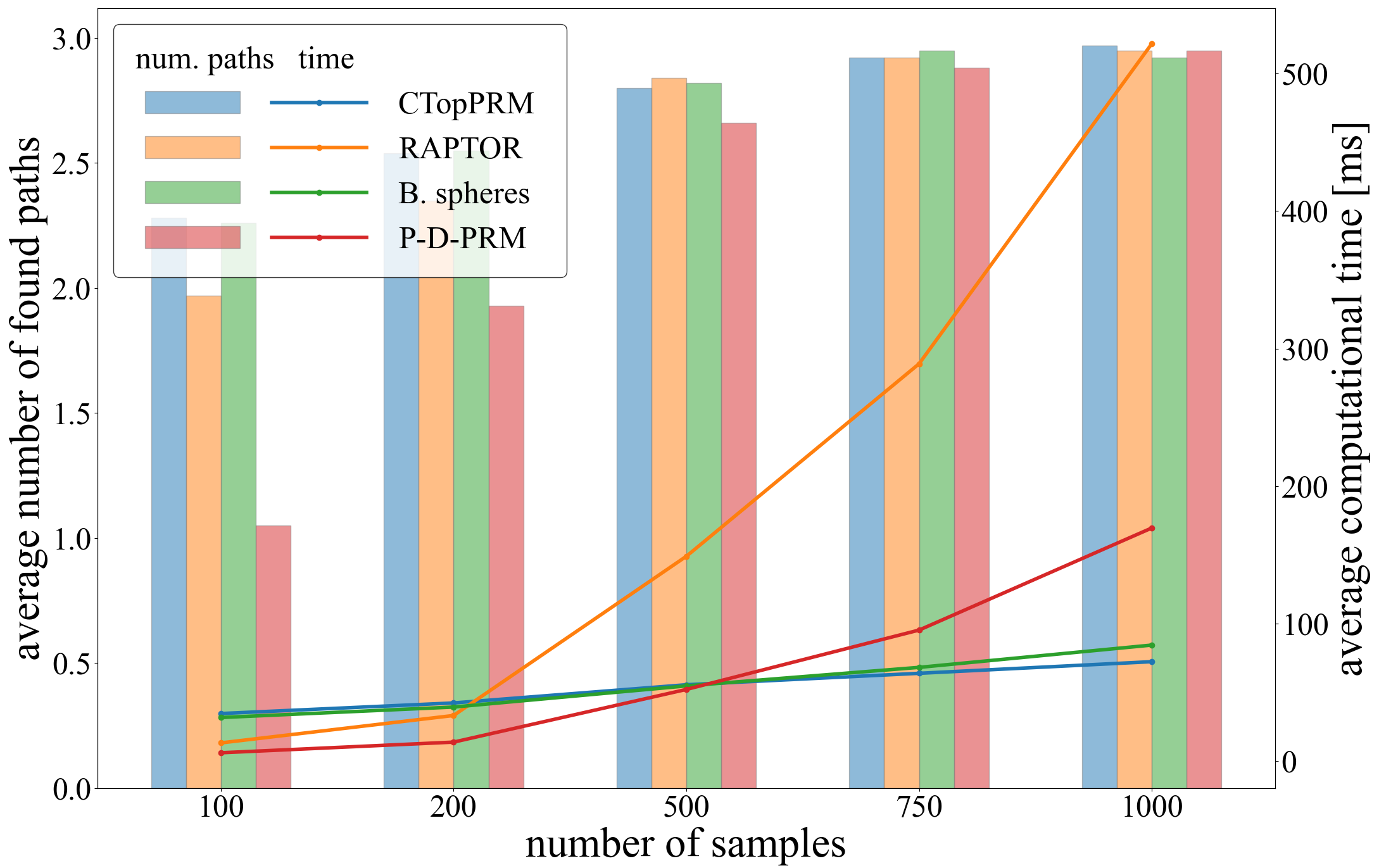}
     \caption{%
     The performance with  the increasing number of random samples.
     }
     \label{fig:samples}
     \vspace{-1.0em}
\end{figure}

%% file: tab/all_maps.tex
\begin{table*}[!htb] 
    \vspace{0.5em}
   \centering 
   \footnotesize 
   {\renewcommand{\tabcolsep}{0.95pt} 
   \caption{Quantity and quality of paths found in poles and building environments\label{tab:all}} 
   \vspace{-1em} 
   \begin{tabular}{r|c|cccc|cccc|cccc|cccc}
     \toprule 
    & & \multicolumn4{c|}{\textbf{CTopPRM}} & \multicolumn4{c|}{RAPTOR\cite{zhou2021raptor}} & \multicolumn4{c|}{B. spheres\cite{penicka2022quadrotor}} & \multicolumn4{c}{P-D-PRM\cite{simeon2008deformation}} \\ 
    map & $n$ & c.t.[\si{ms}] & best & average & n-shortest & c.t.[\si{ms}] & best & average & n-shortest & c.t.[\si{ms}] & best & average & n-shortest & c.t.[\si{ms}] & best & average & n-shortest\\ 
       \midrule 
&400 &19 &\textbf{19} &\textbf{15.16}$\pm$3.76 &7.83 &14 &\textbf{19} &13.71$\pm$3.03 &\textbf{7.75} &\textbf{10} &11 &7.47$\pm$1.48 &7.87 &55 &\textbf{19} &14.11$\pm$1.82 &7.80 \\ 
poles&400 &16 &\textbf{6} &\textbf{4.80}$\pm$0.96 &\textbf{7.79} &20 &5 &3.61$\pm$0.94 &N/A &\textbf{14} &\textbf{6} &3.42$\pm$1.05 &N/A &135 &\textbf{6} &4.42$\pm$1.08 &\textbf{7.79} \\ 
&400 &13 &\textbf{11} &\textbf{7.85}$\pm$1.71 &\textbf{7.52} &\textbf{7} &9 &7.28$\pm$1.44 &7.54 &13 &8 &4.06$\pm$1.08 &7.60 &27 &8 &7.18$\pm$0.84 &\textbf{7.52} \\ 
\midrule
&300 &142 &\textbf{25} &\textbf{7.78}$\pm$4.43 &\textbf{39.49} &\textbf{29} &2 &0.41$\pm$0.58 &N/A &93 &7 &3.24$\pm$1.18 &46.95 &39 &1 &0.16$\pm$0.37 &N/A \\ 
building&300 &151 &\textbf{36} &\textbf{8.74}$\pm$5.79 &\textbf{35.93} &\textbf{21} &1 &0.01$\pm$0.10 &N/A &114 &10 &2.85$\pm$2.02 &N/A &35 &0 &0.00$\pm$0.00 &N/A \\ 
&300 &124 &\textbf{11} &\textbf{4.77}$\pm$1.74 &\textbf{33.16} &\textbf{29} &1 &0.06$\pm$0.24 &N/A &122 &7 &2.26$\pm$1.17 &N/A &42 &1 &0.02$\pm$0.14 &N/A \\ 
\bottomrule
   \end{tabular} 
   } 
   \vspace{-2em}
\end{table*}